\documentclass[11pt]{article}

\usepackage[preprint]{acl}

\usepackage{times}
\usepackage{latexsym}
 
\usepackage[T1]{fontenc}

\usepackage[utf8]{inputenc}

\usepackage{microtype}

\usepackage{inconsolata}

\usepackage{graphicx}

\usepackage{multirow}
\usepackage[table]{xcolor}
\usepackage{booktabs}
\usepackage{amsmath}
\usepackage{amssymb}
\usepackage{color}
\definecolor{keywordcolor}{rgb}{0.7, 0.1, 0.1}   
\definecolor{commentcolor}{rgb}{0.4, 0.4, 0.4}   
\definecolor{symbolcolor}{rgb}{0.0, 0.1, 0.6}    
\definecolor{sortcolor}{rgb}{0.1, 0.5, 0.1}      
\definecolor{errorcolor}{rgb}{1, 0, 0}           
\definecolor{stringcolor}{rgb}{0.5, 0.3, 0.2}    
\usepackage{listings}

\newcommand{\lean}[1]{\lstinline[language=lean, mathescape=true]{#1}}
\usepackage{tcolorbox}
\definecolor{paperblue}{HTML}{02afef}

\usepackage{xspace}
\newcommand{\verina}{\textsc{Verina}\xspace}
\newcommand{\verinaplus}{\textsc{VerinaPlus}\xspace}
\newcommand{\verinalite}{\textsc{VerinaLite}\xspace}
\newcommand{\gain}[1]{\textcolor{paperblue}{\scriptsize~(#1)}}
\newcommand{\Oo}{\mathcal{O}}
\newcommand{\Tt}{\mathcal{T}}

\newcommand{\Dd}{\mathcal{D}}
\newcommand{\ZZ}{\mathbb{Z}}
\newcommand{\NN}{\mathbb{N}}
\newcommand{\smallsec}[1]{\paragraph{#1.}}
\usepackage{enumitem}
\tcbuselibrary{breakable}
\usepackage{fontawesome}

\title{VeriScale: Adversarial Test-Suite Scaling for Verifiable Code Generation}

\author{
 \textbf{Yifan Bai\textsuperscript{1}}\footnotemark[1],
 \textbf{Xiaoyang Liu\textsuperscript{1}}\thanks{Equal contribution},
 \textbf{Zihao Mou\textsuperscript{2}},
 \textbf{Guihong Wang\textsuperscript{1}},
 \textbf{Jian Yu\textsuperscript{3}}, \\
 \textbf{Shuhan Xie\textsuperscript{4}},
 \textbf{Yantao Li\textsuperscript{5}},
 \textbf{Yangyu Zhang\textsuperscript{6}},
 \textbf{Jingwei Liang\textsuperscript{1,7}}\thanks{Corresponding author: jingwei.liang@sjtu.edu.cn, luotao41@sjtu.edu.cn},
 \textbf{Tao Luo\textsuperscript{1,7,8}}\footnotemark[2]\\
 \textsuperscript{1}School of Mathematical Sciences, Shanghai Jiao Tong University\\
 \textsuperscript{2}School of Science and Engineering, The Chinese University of Hong Kong, Shenzhen\\
 \textsuperscript{3}School of Mathematics, Jilin University
 \textsuperscript{4}School of Mathematical Sciences, Tongji University\\
 \textsuperscript{5}Zhiyuan College, Shanghai Jiao Tong University\\
 \textsuperscript{6}School of Future Technology, South China University of Technology\\
 \textsuperscript{7}Institute of Natural Sciences, Shanghai Jiao Tong University\\
 \textsuperscript{8}MOE-LSC, CMA-Shanghai, Shanghai Jiao Tong University
}

\begin{document}
\maketitle

\begin{abstract}
As large language models (LLMs) are increasingly deployed for software engineering, constructing high-quality benchmarks is crucial for evaluating not just the functional correctness, but also the formal verifiability of generated code. 
However, existing benchmarks are limited by the quantity and quality of positive and negative test cases, leading to an overestimation of model capabilities in generating specifications and implementations.
To address this, we propose VeriScale, a novel framework driven by the adversarial implementations. 
It consists of two stages: test-suite expansion to construct diverse and challenging test cases, and test-suite reduction to distill them into compact yet discriminative suites. 
While VeriScale is general, we instantiate it on \verina to construct \verinaplus, which expands the original test suites by over 83$\times$, and \verinalite, a lightweight 14$\times$ variant.
Our experiments across eight state-of-the-art LLMs demonstrate that \verinaplus exposes substantial model weaknesses hidden by the original benchmark, evidenced by sharp score drops on both SpecGen and CodeGen tasks, whereas \verinalite maintains this discriminative power at a fraction of the evaluation cost.
The enhanced benchmarks and source code are publicly available at 
\url{https://github.com/XiaoyangLiu-sjtu/VeriScale}.
\end{abstract}

\begin{figure*}[t]
    \centering
    \includegraphics{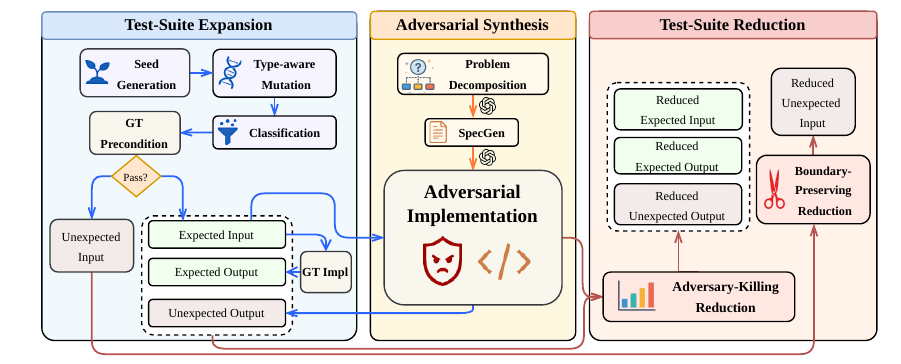}
    \caption{\textbf{Overview of the VeriScale framework for adversarial test-suite scaling.} Driven by the adversarial implementations, the framework scales verifiable code generation benchmarks through two core stages: test-suite expansion to ensure rigorous boundary coverage, and test-suite reduction to optimize evaluation efficiency without sacrificing discriminative power.} 
    \label{fig:overview_framework}
\end{figure*}

\section{Introduction}
The rapid advancement of large language models (LLMs) has reshaped software engineering, particularly in program synthesis and code generation~\citep{chen2021evaluatinglargelanguagemodels, openai2024gpt4technicalreport}. 
However, LLM-generated programs remain prone to subtle logical errors~\citep{asleep}, limiting their deployment in high-stakes settings. 
To address this reliability challenge, verifiable code generation has emerged as a principled paradigm in which models are required to jointly synthesize formal specifications, code implementations, and machine-checkable proofs to establish their consistency~\citep{Clover,ye2026verina}. 
By integrating machine-checkable correctness guarantees into the generation process, this paradigm offers a promising path toward more trustworthy LLM-based software development.

To rigorously evaluate this emerging paradigm, recent benchmarks such as \textsc{Clever}~\citep{thakur2026clever} and \verina~\citep{ye2026verina} have adopted Lean~\citep{moura2021lean} as their core verification backend. 
Unlike verification frameworks such as Dafny~\citep{leino2010dafny} and Verus~\citep{verus}, which heavily leverage SMT solvers~\citep{z3} for automated theorem proving with less explicit proof steps, Lean operates as an interactive theorem prover (ITP). 
This distinction enables the construction of proofs with explicit intermediate steps, thereby facilitating the direct assessment of model capabilities in proof generation (ProofGen). 
For specification generation (SpecGen) and code generation (CodeGen), these benchmarks primarily rely on test cases as the empirical evaluation mechanism~\citep{chen2021evaluatinglargelanguagemodels, ye2026verina}.

However, this reliance on test-suite evaluation exposes a central limitation: the reliability of the assessment depends heavily on the quantity and quality of the underlying test cases.
In current verifiable code generation benchmarks, each problem is typically accompanied by only a limited number of \textbf{positive test cases}, i.e., \textbf{expected input-output pairs}. More importantly, high-quality \textbf{negative test cases} are even more limited, including \textbf{unexpected inputs and unexpected outputs} that are essential for testing whether generated specifications correctly reject invalid behaviors.
As a result, models may satisfy the sparse test cases without capturing the program intent, allowing incorrect implementations or unsound and incomplete specifications to achieve inflated pass rates and leading to an overestimate of model capabilities.

To bridge this gap, we propose VeriScale, an adversarial framework that systematically scales test-suite evaluation for verifiable code generation. 
VeriScale first combines LLM-based seed generation with type-aware mutation to construct a large pool of 
candidate inputs. 
These inputs are then classified as either expected or unexpected according to the ground-truth preconditions. 
For the former, VeriScale executes the reference implementation to derive the corresponding outputs, thereby forming robust expected input-output pairs.

\textbf{The core innovation of VeriScale lies in constructing adversarial implementations to drive both the construction of unexpected outputs and the reduction of expected test cases.} 
Instead of directly prompting LLMs to generate arbitrary unexpected outputs, VeriScale synthesizes adversarial implementations designed to exploit weaknesses in LLM-generated specifications, executing them on expected inputs. 
Outputs are retained only when they differ from the reference outputs while being falsely accepted by the generated postconditions, ensuring they are genuinely informative for exposing specification flaws.
Finally, VeriScale repurposes these implementations in a reduction stage, distilling the expanded cases into compact yet discriminative test suites through boundary-preserving and adversary-killing reduction.

To evaluate our approach, we apply VeriScale to \verina, resulting in the \verinaplus and \verinalite benchmarks.
The empirical results underscore the necessity of adversarial test scaling. 
Notably, GPT-5.5 experiences a severe performance degradation on \verinaplus, plummeting from 68.78\% to 44.44\% on SpecGen and from 96.83\% to 86.24\% on CodeGen. 
This stark contrast highlights how the original benchmark significantly overestimates true model capabilities. 
Furthermore, model performance on \verinalite closely mirrors that on \verinaplus, confirming that our reduction strategy successfully eliminates redundant test cases while preserving discriminative rigor.

Our main contributions are as follows:
\begin{itemize}[topsep=0pt, itemsep=0pt, parsep=0pt]
    \item[1.] We propose VeriScale, a framework that leverages adversarial implementations to drive both the systematic expansion and the reduction of test suites for verifiable code generation.
    \item[2.] We instantiate VeriScale on \verina to construct \verinaplus, expanding the original test suites by over $83\times$, alongside \verinalite, a lightweight $14\times$ variant.
    \item[3.] We show that \verinaplus reveals substantial weaknesses overlooked by the original benchmark in both SpecGen and CodeGen, while \verinalite retains the discriminative power at a fraction of the evaluation cost.
\end{itemize}

\section{Related Work}
\smallsec{Benchmarks for Verifiable Code Generation}
The evaluation of LLM-generated programs has evolved from functional correctness toward formal verifiability. 
Early benchmarks such as HumanEval~\citep{chen2021evaluatinglargelanguagemodels}, MBPP~\citep{austin2021programsynthesislargelanguage}, APPS~\citep{hendrycks2021measuring}, and LiveCodeBench~\citep{jain2025livecodebench} primarily assess code generation through unit tests, but do not evaluate formal specifications or proofs. 
Recent verification-oriented benchmarks, including Dafny-Synthesis~\citep{DafnySynthesis}, Clover~\citep{Clover}, DafnyBench~\citep{loughridge2025dafnybench}, miniCodeProps~\citep{lohn2024minicodeprops}, FVAPPS~\citep{dougherty2025provingcodinginterviewbenchmark}, and \verina~\citep{ye2026verina}, incorporate formal reasoning tasks in languages such as Dafny and Lean.

Among these benchmarks, \verina provides a comprehensive benchmark for verifiable code generation, supporting the evaluation of SpecGen, CodeGen, and ProofGen.
VeriScale is complementary to these efforts: instead of introducing new tasks, it strengthens existing benchmarks by scaling expected input-output pairs, unexpected inputs, and high-quality unexpected outputs for more discriminative test-suite evaluation.

\smallsec{Scaling Test-Suite Evaluation}
Test-suite quality is crucial for reliable code generation evaluation, as limited unit tests may allow incorrect programs to pass by missing corner cases. 
EvalPlus~\citep{evalplus} is a representative effort that augments HumanEval~\citep{chen2021evaluatinglargelanguagemodels} with large-scale test cases by combining seed generation and type-aware mutation. 
Mutation-based generation can mitigate benchmark leakage, since newly generated tests are less likely to have been memorized during training~\citep{xu2024benchmarkingbenchmarkleakagelarge, 2026memorizationinLLM}. 
Adversarial tests have also proven effective in exposing weaknesses in code generation and program repair systems~\citep{CarlAdversarialAttacks, AdverIntentAgent}.

VeriScale is inspired by these methods but targets verifiable code generation, where evaluation requires richer signals than expected input-output pairs alone. 
In particular, SpecGen evaluation needs unexpected inputs and unexpected outputs to test whether generated specifications correctly reject invalid behaviors. 
Therefore, VeriScale extends test-suite scaling with precondition-guided classification and adversarial implementation synthesis, producing more informative negative cases to strictly evaluate the SpecGen task.

\section{Methodology}
In this section, we detail VeriScale, a framework designed to scale test-suite evaluation for verifiable code generation, as illustrated in Figure~\ref{fig:overview_framework}. 
The framework comprises two primary stages, test-suite expansion and test-suite reduction, both fundamentally driven by adversarial implementations. 

\subsection{Test-Suite Expansion}
Formally, each benchmark task is equipped with a problem description, a ground-truth precondition, a reference implementation, and existing base inputs. 
From these components, \textbf{VeriScale constructs a comprehensive test suite encompassing three types of cases: expected input-output pairs, unexpected inputs, and unexpected outputs}. 
Expected input-output pairs support both CodeGen and SpecGen evaluation by checking intended behaviors, while unexpected inputs and unexpected outputs test whether generated specifications reject invalid inputs and incorrect outputs, respectively.

\smallsec{Seed Generation and Type-aware Mutation}
Drawing inspiration from EvalPlus~\citep{evalplus}, VeriScale employs LLMs to synthesize a diverse set of initial seed inputs. 
To construct the generation prompt for each task, we aggregate the problem description, the ground-truth precondition, input parameter signatures (names and types), and any pre-existing expected and unexpected inputs. 
The LLM is then queried to output a specified volume of distinct candidate inputs, establishing a seed pool for the subsequent type-aware mutation phase. 
The detailed prompt is shown in Appendix~\ref{sec:prompts}.

To systematically expand the initial seed pool, VeriScale employs an iterative, type-aware mutation strategy~\citep{park2021generative}.
To preserve task-specific semantics during structural perturbation, we extract an ingredient pool from existing candidates, allowing target parameters to stochastically reuse these contextual values \citep{martinez2019astor}.
When reuse is not triggered, the framework falls back on type-specific heuristic perturbations. 
To construct complex edge cases, the mutation engine performs a sequence of modifications by randomly targeting a single parameter per step under a bounded generation budget. 
To mathematically capture how a specific value is perturbed during any given step, we formalize our type-aware mutation procedure as a five-tuple $\mathcal{M} = (\mathcal{T}, \mathcal{D}, \mathcal{O}, p, q)$.
\begin{itemize}[itemsep=0pt, parsep=0pt]
    \item $\Tt$ is the set of input types.
    \item $\Dd$ maps each type $\tau \in \Tt$ to its domain $D_\tau$.
    \item  $\Oo$ maps each type $\tau \in \Tt$ to a finite set of mutation schemas $\Oo_\tau = \{s_{\tau,1}, \ldots, s_{\tau,k_\tau}\}$. Each schema $s_{\tau,i}: D_{\tau}\times \Theta_{\tau,i} \to D_{\tau}$ is parameterized by a parameter space $\Theta_{\tau,i}$.
    \item $p$ maps each type $\tau \in \mathcal{T}$ to a probability distribution $p_\tau$ over $\mathcal{O}_\tau$, specifying how a mutation schema is selected.
    \item $q$ maps each pair $(\tau,i)$ to a probability distribution $q_{\tau,i}$ over the parameter space $\Theta_{\tau,i}$, specifying how the parameters of schema $s_{\tau,i}$ are sampled. In general, this distribution may depend on the input value $x \in D_\tau$.
\end{itemize}

Given an input value $x \in D_\tau$, the mutator first samples a mutation schema $I \sim p_{\tau}$, and then samples its parameters according to $\theta \sim q_{\tau,I}(\cdot \mid x)$. The mutated value is then defined as
\[
M_\tau(x) = s_{\tau,I}(x;\theta).
\]
Since each schema is type-preserving, for any $x \in D_\tau, i \in \{1,\ldots,k_\tau\}, \theta \in \Theta_{\tau,i}$,  we have
\[
s_{\tau,i}(x;\theta) \in D_\tau.
\]
Therefore, the mutation procedure preserves the input type by construction:
\[
M_\tau(x) \in D_\tau.
\]
See Appendix~\ref{sec:mutation} for more details of these schemas.

\smallsec{Precondition-Guided Input Classification}
To classify the candidate inputs, VeriScale operates in alignment with Verina~\citep{ye2026verina}, evaluating them against the ground-truth precondition using a two-stage verification pipeline.
Prior to logical evaluation, we enforce a strict syntax filter. 
Each candidate is rendered as a Lean precondition application, denoted as \lean{<expr>}, and validated via the \lean{\#check <expr>} command. 
Candidates failing this step due to missing arguments, type mismatches, or syntax errors are immediately discarded.

In the first stage, we apply a bidirectional procedure, evaluating both \lean{\#guard decide <expr>} and \lean{\#guard decide (¬ <expr>)}. 
An input is classified as expected if the positive expression evaluates to true (or the negation to false), and unexpected if the inverse holds. 
For these unresolved inputs, we construct an \lean{example}, unfold the precondition, aggressively simplify the goal context via \lean{simp_all!}, and execute \lean{plausible} bidirectionally. 
Discovering a counterexample for the negated expression confirms the input as expected, whereas finding one for the positive expression marks it as unexpected. 
Candidates that still lack a definitive verdict due to timeouts or search exhaustion are permanently excluded from the final test suite.

Figure~\ref{fig:classification_example} illustrates this bidirectional pipeline in practice. 
For the isolated expected inputs, we execute the reference implementation to compute their corresponding outputs. 
This completes the expected input-output pairs, establishing a ground truth for evaluating functional correctness.

\begin{figure}[ht!]
    \centering
    \begin{tcolorbox}[
        colback=gray!5!white,
        colframe=black!70!white,
        title=\textbf{\textsc{Example of Candidate Input Classification}},
        fonttitle=\bfseries\small,
        boxrule=0.6pt,
        arc=1.5mm,
        left=2mm, right=2mm, top=2mm, bottom=2mm
    ]
    
    \small
    \textbf{Problem Description:} (\verina \#advanced\_7) This task requires writing a Lean 4 function that converts a binary number represented as a list of digits (0 or 1) into its corresponding decimal value. The list is ordered in big-endian format, meaning the most significant digit comes first. The function should interpret the list as a binary number and return its decimal representation as a natural number. \\
    \textbf{Ground-Truth Precondition:} \\
    \lean{def binaryToDecimal_precond (digits : List Nat) : Prop :=
  digits.all (fun d => d = 0 ∨ d = 1)} \\
    \vspace{-0.32cm}
    \par\noindent\hrulefill
    \vspace{0.15cm}
    
    \small
    \textbf{Candidate Input:} \lean{[1, 2, 1]}
    \vspace{0.1cm}
    
    \textbf{Syntax Filter:} \lean{\#check binaryToDecimal_precond ([1, 2, 1])}~~\textcolor{blue}{\faCheck~Passed}
    \vspace{0.1cm}
    
    \textbf{Stage 1: Bidirectional \lean{decide}}
    \begin{itemize}[leftmargin=*, nosep]
        \item \lean{\#guard decide (binaryToDecimal_precond ([1, 2, 1]))} $\rightarrow$ \textcolor{orange}{Failed (Expression did not evaluate to `true')}
        \item \lean{\#guard decide (¬ binaryToDecimal_precond ([1, 2, 1]))}~~\textcolor{blue}{\faCheck~Passed}
    \end{itemize}
    \vspace{0.1cm}
    
    \textbf{Stage 2: Bidirectional \texttt{plausible}}
    \begin{itemize}[leftmargin=*, nosep]
        \item 
\begin{lstlisting}[language = lean]
example: binaryToDecimal_precond 
 ([1, 2, 1]) := by
  unfold binaryToDecimal_precond
  simp_all!
  plausible   
\end{lstlisting}
$\rightarrow$ \textcolor{orange}{Found a counter-example!}
        \item 
\begin{lstlisting}[language = lean]
example: ¬ binaryToDecimal_precond 
 ([1, 2, 1]) := by
  unfold binaryToDecimal_precond
  simp_all!  
\end{lstlisting}
\textcolor{blue}{\faCheck~Passed}
    \end{itemize}
    \par\noindent\hrulefill
    \vspace{0.15cm}
    
    \textbf{Final Verdict: Unexpected Input}.
    \end{tcolorbox}
    \vspace{-0.2cm}
    \caption{\textbf{A full trajectory of candidate input classification.} The input \lean{[1,2,1]} will be classified as unexpected because \lean{\#guard descide} fails at Stage 1. According to our judging process, this example does not proceed to Stage 2; it is shown here for illustration only.}
    \label{fig:classification_example}
\end{figure}

\begin{figure*}[t]
    \centering
    \includegraphics[scale = 0.9]{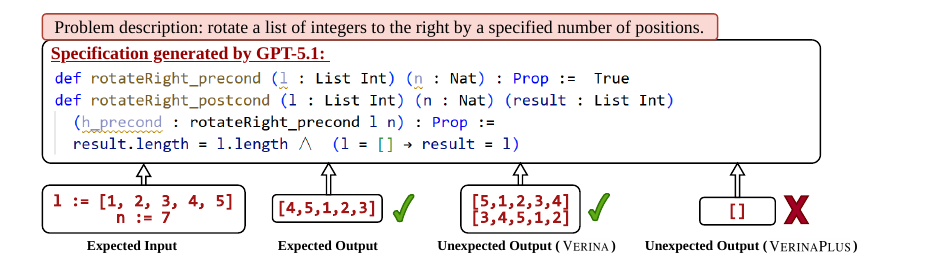}
    \vspace{-0.5cm}
    \caption{\textbf{Case study of a flawed specification on \verina \#advanced\_3.} The generated postcondition successfully rejects unexpected outputs from the original dataset, but erroneously accepts our adversarially synthesized unexpected outputs.}
    \label{fig:unexpected_output_example}
\end{figure*}

\smallsec{Adversarial Synthesis of Unexpected Outputs}
Beyond verifying basic functionality, evaluating the soundness and completeness of formal specifications requires high-quality unexpected outputs to serve as negative test cases. 
Existing approaches often rely on LLMs to directly hallucinate incorrect outputs or employ naive random sampling to deviate from the ground truth. 
However, these trivially incorrect outputs are usually easily rejected by even highly flawed specifications, failing to provide meaningful diagnostic feedback for underspecification.
As illustrated in Figure~\ref{fig:unexpected_output_example}, a concrete case study demonstrates this limitation.

To synthesize highly discriminative unexpected outputs, VeriScale introduces an adversarial execution paradigm. 
We first prompt language models to perform problem decomposition to reduce task complexity and clarify structural logic~\citep{dsr}.
Next, we leverage advanced models to generate high-quality formal specifications.
Crucially, a more capable model acts as a red team~\citep{perez-etal-2022-red}, crafting adversarial implementations to exploit vulnerabilities within these specifications.
As a fallback, if the model fails to generate an adversarial implementation, we systematically drop constraints from the ground-truth specification to expose an attack target, thereby facilitating the successful synthesis of an adversarial implementation.

These adversarial implementations are subsequently executed on the expected inputs generated during the previous stage. 
If an adversarial implementation produces an output that diverges from the ground truth but is still successfully accepted by the generated postconditions, this mismatched value is formally retained. 
Because these unexpected outputs stem from systematic specification gaming rather than arbitrary hallucination, they closely mimic realistic logical errors, making them significantly more effective at exposing specification unsoundness and incompleteness.

    
    


    

\subsection{Test-Suite Reduction}
To reduce computational overhead, we filter unexpected inputs via boundary-preserving reduction and eliminate redundant expected input-output pairs through adversary-killing reduction.
Unexpected outputs are implicitly filtered alongside their corresponding inputs.

\smallsec{Boundary-Preserving Reduction}
For unexpected inputs, we adopt a boundary-preserving reduction strategy rather than relying solely on random sampling or feature diversity. 
We extract lightweight structural signatures (e.g., empty containers, zero or negative values, mismatched lengths, and ordering patterns) to approximate failure modes.
We then prioritize inputs that correspond to common boundary violations. 
For each task, we first retain representative examples from these critical boundary buckets, ensuring that frequent and semantically meaningful unexpected inputs are preserved. 
If the budget is not exhausted, we further fill the remaining slots with representative inputs from diverse structural buckets, and finally use a deterministic priority order as a fallback. 
This approach keeps the unexpected input set compact while maintaining coverage of typical precondition violations and edge-case behaviors.

\begin{table*}[t]
\centering
\small
\setlength{\tabcolsep}{7pt}
\renewcommand{\arraystretch}{0.95}
\caption{\textbf{Comparison of test suite volumes.} Data is formatted as Mean~(Min--Max), accompanied by the relative growth multiplier compared to the baseline \verina.}
\label{tab:dataset-stats}
\begin{tabular}{lccc}
\toprule[1pt]
\textbf{Dataset} 
& \textbf{Expected Input-Output} 
& \textbf{Unexpected Output} 
& \textbf{Unexpected Input} \\
\midrule[0.1pt]
\verina & 5.89~(2--13) & 12.69~(2--34) & 0.65~(0--7) \\
\verinaplus & 370.07~(5--705)~\gain{$\times 62.83$} & 1114.01~(15--2973)~\gain{$\times 87.79$} & 119.00~(0--684)~\gain{$\times 183.08$} \\
\verinalite & 52.34~(5--61)~\gain{$\times 8.89$} & 202.35~(15--467)~\gain{$\times 15.95$} & 15.80~(0--50)~\gain{$\times 24.31$} \\
\bottomrule[1pt]
\end{tabular}
\end{table*}

\smallsec{Adversary-Killing Reduction}
We further use adversary-killing to reduce the set of expected input-output pairs.
Specifically, we repurpose the adversarial implementations synthesized in the previous stage, treating each as a distinct mutant.
We associate each expected pair with the set of adversarial implementations it can kill, formulating test reduction as a set-cover problem.
We first select expected pairs that cover the largest number of previously unkilled mutants using a greedy procedure, ensuring that every implementation detectable by the expanded test suite remains covered by at least one retained pair. 
Then, under a fixed per-task budget, we add additional pairs that kill the highest number of implementations. 
This strategy avoids the coverage loss caused by purely diversity-based or random reduction, allowing the lightweight test suite to substantially reduce its size while preserving its ability to distinguish flawed specifications.

\begin{figure}[ht!]
    \centering
    \begin{tcolorbox}[
        colback=gray!5!white,
        colframe=black!70!white,
        title=\textbf{\textsc{Example of Adversary-Killing Reduction}},
        fonttitle=\bfseries\small,
        boxrule=0.6pt,
        arc=1.5mm,
        left=2mm, right=2mm, top=2mm, bottom=2mm
    ]
    
    \small
    \textbf{Problem Description:} (\verina \#advanced\_16) Implement the insertion sort algorithm in Lean 4. The function takes a single list of integers as input and returns a new list that contains the same integers in ascending order.\\
    \vspace{-0.32cm}
    \par\noindent\hrulefill
    \vspace{0.15cm}
    
    \small
    \textbf{Candidate Expected Input:} \lean{[0, -1, -2, -3, -4]}
    \vspace{0.1cm}
    
    \textbf{Expected Output:} \lean{[-4, -3, -2, -1, 0]}
    \vspace{0.1cm}
    
    
    \textbf{Adversary-Killing Count}
    \begin{itemize}[leftmargin=*, nosep]
        \item 
\begin{lstlisting}[language = lean]
def insertionSort (xs : List Int) : 
 List Int := xs.reverse
\end{lstlisting}
  \textbf{Output:} \lean{[-4, -3, -2, -1, 0]} $\rightarrow$ \textcolor{orange}{Same as expected}
        \item 
\begin{lstlisting}[language = lean]
def insertionSort (xs : List Int) :
 List Int := match xs with
  | [] => []
  | x :: _ => List.replicate xs.length x   
\end{lstlisting}
    \textbf{Output:} \lean{[0, 0, 0, 0, 0]}~~\textcolor{blue}{\faCheck~Different}
    \item
\begin{lstlisting}[language = lean]
def insertionSort (xs : List Int) : 
 List Int := List.replicate xs.length 0
\end{lstlisting}
    \textbf{Output:} \lean{[0, 0, 0, 0, 0]}~~\textcolor{blue}{\faCheck~Different}
    \item 
\begin{lstlisting}[language = lean]
def insertionSort (xs : List Int) : 
 List Int := (List.range xs.length).map (fun n => Int.ofNat n)
\end{lstlisting}
    \textbf{Output:} \lean{[0, 1, 2, 3, 4]}~~\textcolor{blue}{\faCheck~Different}
    \item 
\begin{lstlisting}[language = lean]
def insertionSort (xs : List Int) : 
 List Int := []
\end{lstlisting}
    \textbf{Output:} \lean{[]}~~\textcolor{blue}{\faCheck~Different}
    \item 
\begin{lstlisting}[language = lean]
def insertionSort (xs : List Int) : 
 List Int := xs
\end{lstlisting}
    \textbf{Output:} \lean{[0, -1, -2, -3, -4]}~~\textcolor{blue}{\faCheck~Different}
    \item
\begin{lstlisting}[language = lean]
def insertionSort (xs : List Int) : List Int :=
  match xs with
  | [] => []
  | x :: _ => [x]
\end{lstlisting} 
\textbf{Output:} \lean{[0]}~~\textcolor{blue}{\faCheck~Different}
    \end{itemize}

    \par\noindent\hrulefill
    \vspace{0.15cm}
    
    \textbf{Adversary-Killing Count: 6 / 7}.
    \end{tcolorbox}
    \vspace{-0.2cm}
    \caption{\textbf{An example of adversary-killing reduction.} Given an expected input, an adversarial implementation is considered "killed" if its output differs from the expected output. The illustrated case kills six out of seven adversarial implementations, making it a high-quality case. For simplicity, we ignore any preconditions in these adversarial implementations.}
    \label{fig:adversarial_kill_example}
\end{figure}

\section{Experiments}
\subsection{The Augmented Benchmarks}
We instantiate VeriScale on the \verina dataset~\citep{ye2026verina} to construct the augmented benchmarks \verinaplus and \verinalite. 
During the expansion stage, we employ GPT-5.3-Codex for seed generation, problem decomposition, and adversarial implementation synthesis. 
To obtain highly diverse candidate specifications, we utilize an array of widely adopted prior models: Claude-Haiku-4.5, Claude-Sonnet-4.5, GPT-5.1, GPT-4.1, Qwen3-Max, and DeepSeek-V3.2.
Detailed hyperparameter configurations of the expansion and reduction stages are provided in Appendix~\ref{sec:hyperparameter}.
In total, the API cost incurred during the benchmark expansion phase was \$58.89.

As demonstrated in Table~\ref{tab:dataset-stats}, \verinaplus significantly scales up the test volume by over 83$\times$ compared to the original baseline to ensure rigorous boundary coverage, whereas \verinalite optimizes computational overhead via the aforementioned reduction strategies to provide a lightweight 14$\times$ variant. 
Notably, the augmented benchmarks also consistently maintain 100\% code coverage.

\subsection{Evaluation Setup}
We evaluate eight state-of-the-art LLMs, including Claude-Sonnet-4.6, Claude-Opus-4.7, GPT-5.3-Codex, GPT-5.5, Gemini-3.1-Flash-Lite-Preview, Gemini-3.1-Pro-Preview, Qwen3.6-Max-Preview, and DeepSeek-V4-Pro. 
These models are assessed across the \verina, \verinaplus, and \verinalite benchmarks, focusing on the SpecGen and CodeGen tasks. 
To ensure a rigorous assessment, we employ the official evaluation harness provided by \verina~\citep{ye2026verina} to compute and report the pass@1 scores. 
However, we intentionally remove the original 2-shot setting to facilitate the evaluation of the entire dataset.

\begin{figure*}[t]
    \centering
    \includegraphics[width=1\linewidth]{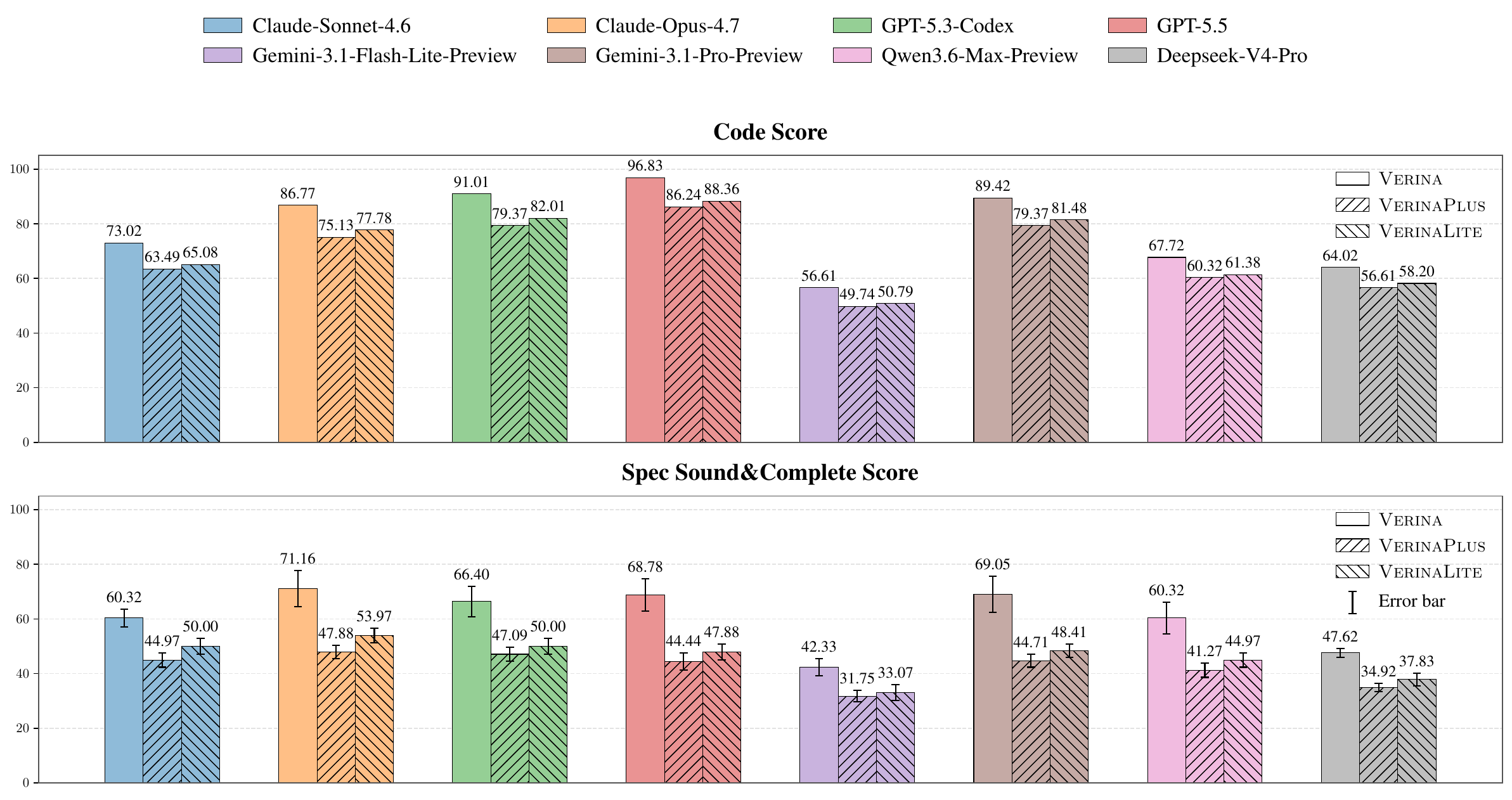}
    \caption{\textbf{Performance comparison of eight models on \verina, \verinaplus, and \verinalite.} The top panel shows the Code Score, while the bottom panel displays the Spec Sound\&Complete Score. For the specification evaluation, we include error bars indicating the lower bound (treating unknown cases as not holding) and upper bound (treating unknown cases as holding).}
    \label{fig:experiment}
\end{figure*}

All experiments are implemented using Lean v4.24.0. 
The evaluation was conducted on a server equipped with dual Intel Xeon Platinum 8380 CPUs, providing a total of 80 physical cores.

\subsection{Evaluation Results}
To systematically interpret our findings, we organize the evaluation results across three critical dimensions: the effectiveness of the augmented benchmarks in exposing hidden model vulnerabilities, the profound robustness gap between SpecGen and CodeGen, and the computational efficiency of our test-suite reduction strategy.

\smallsec{Effectiveness of the Augmented Benchmarks}
As illustrated in Figure~\ref{fig:experiment}, all models exhibit significant performance degradation across both SpecGen and CodeGen tasks on \verinaplus and \verinalite compared to the \verina baseline. 
For instance, the top-performing model, GPT-5.5, achieves a code score of 96.83 and a specification soundness and completeness score of 68.78 on the baseline, which drop sharply to 86.24 and 44.44 on \verinaplus, respectively. 
This consistent decline across all models demonstrates that the original benchmark overestimates model capabilities due to underspecified test cases. 
By addressing this critical limitation, our adversarial expansions successfully systematically expose previously hidden vulnerabilities, providing a much more rigorous evaluation of the SpecGen and CodeGen tasks.

Furthermore, a closer examination of the SpecGen evaluation reveals a striking reduction in evaluation uncertainty, as evidenced by the significantly narrowed error bars on the augmented benchmarks.
In the original \verina baseline, the scarcity of test cases frequently leads to indeterminate verification outcomes, resulting in wide margins between the upper and lower bounds. 
For example, Claude-Opus-4.7 exhibits a substantial gap of 13.23\% (77.78\% vs. 64.55\%) on the baseline. 
However, under the rigorous scrutiny of \verinaplus, this ambiguity is effectively resolved, drastically shrinking the gap for the same model to merely 4.76\% (50.26\% vs. 45.50\%). 
This consistent compression of error margins across all evaluated models demonstrates that our adversarial scaling not only exposes model flaws, but also transforms a previously ambiguous evaluation into a highly definitive and statistically reliable assessment.

\begin{table*}[t]
\centering
\small
\setlength{\tabcolsep}{4pt}
\renewcommand{\arraystretch}{0.95}
\caption{\textbf{Evaluation time on \verina, \verinaplus, and \verinalite.} We report wall-clock time in seconds, with the average row additionally highlighting the relative time multipliers compared to the \verina baseline.}
\label{tab:time}
\begin{tabular}{lcccccc}
\toprule[1pt]
\multirow{2}{*}{\textbf{Model}} 
& \multicolumn{3}{c}{\textbf{SpecGen}} 
& \multicolumn{3}{c}{\textbf{CodeGen}} \\
\cmidrule(lr){2-4} \cmidrule(lr){5-7}
& \textbf{\verina} & \textbf{\verinaplus} & \textbf{\verinalite}
& \textbf{\verina} & \textbf{\verinaplus} & \textbf{\verinalite} \\
\midrule[0.1pt]
Claude-Sonnet-4.6 & 996 & 1772 & 1280 & 268 & 356 & 241 \\
Claude-Opus-4.7 & 1324 & 2758 & 2102 & 250 & 394 & 269 \\
GPT-5.3-Codex & 1018 & 1839 & 1395 & 252 & 413 & 327 \\
GPT-5.5 & 1191 & 2100 & 1669 & 264 & 413 & 284 \\
Gemini-3.1-Flash-Lite-Preview & 790 & 1333 & 1053 & 190 & 305 & 205 \\
Gemini-3.1-Pro-Preview & 1214 & 2081 & 1645 & 254 & 396 & 267 \\
Qwen3.6-Max-Preview & 962 & 1732 & 1333 & 212 & 329 & 227 \\
DeepSeek-V4-Pro & 757 & 1261 & 935 & 260 & 311 & 217 \\
\midrule[0.1pt]
\textbf{Average} & 1032 & 1860~\gain{$\times 1.80$} & 1427~\gain{$\times 1.38$} & 244 & 365~\gain{$\times 1.50$} & 255~\gain{$\times 1.05$} \\
\bottomrule[1pt]
\end{tabular}
\end{table*}

\smallsec{Robustness Gap between SpecGen and CodeGen}
While the augmented benchmarks universally depress scores, this degradation is far from uniform: the decline in CodeGen is drastically outpaced by a precipitous drop in SpecGen.
Taking the top-performing GPT-5.5 as a representative example again, the model experiences a moderate relative decline of roughly 10.59\% in CodeGen when moving from the \verina baseline to \verinaplus, yet suffers a staggering 24.34\% relative reduction in its specification soundness and completeness score.
This consistent disparity, observed across all evaluated LLMs, strongly indicates that synthesizing sound and complete specifications is fundamentally more challenging than generating functional code.

This stark contrast highlights a fundamental limitation in the current state-of-the-art models. 
Thanks to vast exposure to standard software repositories during pre-training, advanced LLMs have developed a remarkable proficiency for synthesizing functional programming logic. 
However, this empirical success is heavily reliant on statistical pattern matching and does not seamlessly transfer to the rigorous domain of formal verification. 
Synthesizing sound and complete specifications demands strict mathematical reasoning and a global awareness of constraints. 
Consequently, when tasked with formalizing properties that defend against adversarial edge cases, the reasoning capabilities of these models remain highly fragile.

\smallsec{Efficiency of Test-Suite Reduction Strategy}
Beyond evaluation rigor, the practical utility of a benchmark heavily depends on its computational overhead. 
As detailed in Table~\ref{tab:time}, exhaustively evaluating models on the expanded \verinaplus incurs a substantial computational penalty due to the sheer volume of test cases. 
Specifically, this exhaustive approach multiplies the average time overhead to 1.80 and 1.50 times the baseline for the SpecGen and CodeGen tasks, respectively. 
Such pronounced time constraints can significantly hinder the widespread utilization of the benchmark.

Crucially, our minimized benchmark, \verinalite, effectively resolves this scalability issue without sacrificing evaluative rigor.
It drastically reduces the time overhead to merely 1.38 and 1.05 times the original baseline. 
More importantly, while achieving this efficiency, \verinalite simultaneously maintains a performance degradation profile highly similar to the full \verinaplus suite (as evidenced in Figure~\ref{fig:experiment}). 
This optimal balance proves that our reduction strategies successfully prune redundant cases while fully retaining the essential diagnostic power, thereby establishing \verinalite as a highly practical benchmark for evaluations of verifiable code generation.

\section{Conclusion and Future Work}
In this paper, we proposed VeriScale, an adversarial framework that systematically expands and reduces test suites to rigorously evaluate verifiable code generation. 
We instantiated VeriScale on the original \verina benchmark to construct \verinaplus (expanded by over 83$\times$) and \verinalite (a 14$\times$ variant).
Our evaluations demonstrate that \verinaplus uncovers substantial weaknesses in state-of-the-art LLMs across both SpecGen and CodeGen tasks that original benchmarks fail to detect, while \verinalite achieves the same discriminative power with significantly reduced evaluation cost. 
Together, these augmented benchmarks provide a crucial, highly reliable standard for advancing LLM-based verifiable code generation.

Future work will evolve VeriScale into a closed-loop repair system, using the discovered counterexamples to guide models in iteratively patching flawed specifications. 
Furthermore, recognizing that SpecGen is fundamentally an autoformalization task severely bottlenecked by data scarcity, we plan to leverage VeriScale as an adversarial data engine. 
By synthesizing large-scale, high-quality training corpora, we aim to train a dedicated model capable of robust autoformalization.
Finally, generalizing VeriScale to other formal verification frameworks, such as Dafny, remains a critical step toward verifiable code generation.

\section*{Limitations}
While VeriScale significantly advances the rigorous evaluation of verifiable code generation, it has several limitations. 
First, the pipeline relies fundamentally on the availability of ground-truth preconditions and the reference implementations.
Second, the quality of the generated unexpected outputs is inherently bounded by the reasoning and formalization capabilities of the LLM acting as the red team. 
If the model struggles with the strict typing constraints of Lean 4, it may fail to synthesize sophisticated adversarial implementations. 
Third, the construction phase of the expanded test suites remains highly compute-intensive, requiring massive LLM sampling and Lean executions.

\section*{Acknowledgments}
This work is sponsored by the National Key R\&D Program of China Grant No. 2022YFA1008200 (T. L.). We also thank Shanghai Institute for Mathematics and Interdisciplinary Sciences (SIMIS) for their financial support. This research was funded by SIMIS under grant number SIMIS-ID-2025-ST. The authors are grateful for the resources and facilities provided by SIMIS, which were essential for the completion of this work.

\bibliography{custom}

\appendix
\onecolumn
\raggedbottom

\section{Detailed Type-aware Mutation Rules}
\label{sec:mutation}

When ingredient pool reuse is not triggered, we apply type-specific mutations for the following Lean input types: \texttt{Int}, \texttt{Nat}, \texttt{List Int}, \texttt{Array Int}, \texttt{List Nat}, \texttt{Array Nat}, and \texttt{List Char}, \texttt{String}.
\smallsec{Integers}
When $\tau =$ \lean{Int}, the domain is $D_{\tau} = \ZZ$. We define mutation-schema set as
\[
\Oo_{\tau} \triangleq \left\{ s_{\tau,1}, s_{\tau,2}\right\}
\]
where
\begin{align*}
    s_{\tau,1}(x,\delta) &= x + \delta, &\Theta_{\tau,1} = \left\{ -2,-1,0,1,2\right\},\\
    s_{\tau,2}(x,\delta) &= \delta x,    &\Theta_{\tau,2} = \left\{ -2,-1,0,1,2\right\}.
\end{align*}
When ingredient pool reuse is disabled, both the mutation schema and its
parameters are sampled uniformly. Therefore,
\begin{align*}
    p_{\tau}(i) &= \frac{1}{2}, \quad i \in \left\{ 1,2 \right\},\\
    q_{\tau,i}(\delta) &= \frac{1}{5}, \quad i \in \left\{ 1,2\right\}, \quad \delta \in \left\{ -2,-1,0,1,2\right\}.
\end{align*}

\smallsec{Natural numbers}
When $\tau = \texttt{Nat}$, the domain is $D_\tau=\NN$.
We obtain the mutation schemas by applying the integer schemas and clipping
the result to the natural-number domain. Formally, let $\operatorname{clip}(z)=\max(0,z)$, we define
\[
\mathcal{O}_{\tau}
\triangleq
\left\{
s_{\tau,1},
s_{\tau,2}
\right\},
\]
where
\begin{align*}
    s_{\tau,1}(x;\delta)
    &= \operatorname{clip}(x+\delta),
    &
    \Theta_{\tau,1}
    &= \{-2,-1,0,1,2\},
    \\
    s_{\tau,2}(x;\delta)
    &= \operatorname{clip}(\delta x),
    &
    \Theta_{\tau,2}
    &= \{-2,-1,0,1,2\}.
\end{align*}
The clipping operation ensures that every mutated value remains in
$D_\tau=\mathbb{N}$. When ingredient pool reuse is disabled, both the mutation
schema and its parameters are sampled uniformly:
\begin{align*}
    p_{\tau}(i)
    &= \frac{1}{2},\quad
    i \in \{1,2\},
    \\
    q_{\tau,i}(\delta)
    &= \frac{1}{5},
    \quad i \in \{1,2\},
    \quad
    \delta \in \{-2,-1,0,1,2\}.
\end{align*}

\smallsec{Integer lists and arrays}
When $\tau\in\{\texttt{List Int}, \texttt{Array Int}\}$, we model both lists and arrays as finite sequences. Thus,
\[
D_\tau = \left( \ZZ^* \triangleq \bigcup_{n \in \NN} \mathbb{Z}^n \right).
\]
We define the mutation-schema set as
\[
\Oo_{\tau}
\triangleq
\left\{
s_{\tau,1},
s_{\tau,2},
s_{\tau,3},
s_{\tau,4}
\right\},
\]
corresponding to element modification, appending, deletion, and reversal.
For a sequence $x$, let $|x|$ denote its length, $x_j$ its $j$-th element,
$\operatorname{replace}(x,j,v)$ the sequence obtained by replacing $x_j$ with
$v$, and $\operatorname{del}(x,j)$ the sequence obtained by deleting $x_j$.
The schemas are defined as
\begin{align*}
    s_{\tau,1}(x;j,i,\delta)
    &=
    \begin{cases}
    \operatorname{replace}\!\left(x,j,s_{\texttt{Int},i}(x_j;\delta)\right),
    & |x|>0,\\
    x, & |x|=0,
    \end{cases}
    \\
    s_{\tau,2}(x;v)
    &= x \mathbin{++} [v],
    \\
    s_{\tau,3}(x;j)
    &=
    \begin{cases}
    \operatorname{del}(x,j), & |x|>0,\\
    x, & |x|=0,
    \end{cases}
    \\
    s_{\tau,4}(x)
    &= \operatorname{rev}(x).
\end{align*}
The corresponding parameter spaces are
\begin{align*}
    \Theta_{\tau,1}(x)
    &=
    \{0,\ldots,|x|-1\}
    \times
    \{1,2\}
    \times
    \{-2,-1,0,1,2\},
    \\
    \Theta_{\tau,2}
    &= \{-5,-4,\ldots,5\},
    \\
    \Theta_{\tau,3}(x)
    &=
    \{0,\ldots,|x|-1\},
    \\
    \Theta_{\tau,4}
    &= \varnothing.
\end{align*}
When $|x|=0$, the index-dependent schemas are interpreted as the identity
function, so the mutation remains well-defined. When ingredient pool reuse is
disabled, the schema distribution is uniform:
\[
p_\tau(i)=\frac{1}{4},
\quad
i\in\{1,2,3,4\}.
\]
The parameter distributions are uniform over their corresponding finite
parameter spaces. In particular,
\begin{align*}
    q_{\tau,1}(j,i,\delta \mid x)
    &=
    \frac{1}{|x|}\cdot \frac{1}{2}\cdot \frac{1}{5},
    &
    j&\in\{0,\ldots,|x|-1\},
    \quad
    i\in\{1,2\},
    \quad
    \delta\in\{-2,-1,0,1,2\},
    \\
    q_{\tau,2}(v)
    &=
    \frac{1}{11},
    &
    v&\in\{-5,-4,\ldots,5\},
    \\
    q_{\tau,3}(j\mid x)
    &=
    \frac{1}{|x|},
    &
    j&\in\{0,\ldots,|x|-1\},
    \\
    q_{\tau,4}(*) &= 1.
\end{align*}

\smallsec{Natural-number lists and arrays}
When $\tau\in\{\texttt{List Nat},\texttt{Array Nat}\}$, domain of the two types are both
\[
D_\tau = \left( \NN^* \triangleq  \bigcup_{n \in \NN} \NN^n \right).
\]
The mutation schemas are analogous to those for integer lists and arrays,
except that every newly generated element is clipped to the natural-number
domain. Let $\operatorname{clip}(z)=\max(0,z)$. We define
\[
\Oo_{\tau}
\triangleq
\left\{
s_{\tau,1},
s_{\tau,2},
s_{\tau,3},
s_{\tau,4}
\right\},
\]
where
\begin{align*}
    s_{\tau,1}(x;j,i,\delta)
    &=
    \begin{cases}
    \operatorname{replace}\!\left(
        x,
        j,
        \operatorname{clip}\!\left(s_{\texttt{Int},i}(x_j;\delta)\right)
    \right),
    & |x|>0,\\
    x, & |x|=0,
    \end{cases}
    \\
    s_{\tau,2}(x;v)
    &= x \mathbin{++} [\operatorname{clip}(v)],
    \\
    s_{\tau,3}(x;j)
    &=
    \begin{cases}
    \operatorname{del}(x,j), & |x|>0,\\
    x, & |x|=0,
    \end{cases}
    \\
    s_{\tau,4}(x;*)
    &= \operatorname{rev}(x).
\end{align*}
The parameter spaces are
\begin{align*}
    \Theta_{\tau,1}(x)
    &=
    \{0,\ldots,|x|-1\}
    \times
    \{1,2\}
    \times
    \{-2,-1,0,1,2\},
    \\
    \Theta_{\tau,2}
    &= \{-5,-4,\ldots,5\},
    \\
    \Theta_{\tau,3}(x)
    &=
    \{0,\ldots,|x|-1\},
    \\
    \Theta_{\tau,4}
    &= \varnothing.
\end{align*}
For empty lists or arrays, we again interpret index-dependent mutations as
identity transformations. When ingredient pool reuse is disabled,
\[
p_\tau(i)=\frac{1}{4},
\quad
i\in\{1,2,3,4\}.
\]
The parameters are sampled uniformly:
\begin{align*}
    q_{\tau,1}(j,i,\delta \mid x)
    &=
    \frac{1}{|x|}\cdot \frac{1}{2}\cdot \frac{1}{5},
    &
    j&\in\{0,\ldots,|x|-1\},
    \quad
    i\in\{1,2\},
    \quad
    \delta\in\{-2,-1,0,1,2\},
    \\
    q_{\tau,2}(v)
    &=
    \frac{1}{11},
    &
    v&\in\{-5,-4,\ldots,5\},
    \\
    q_{\tau,3}(j\mid x)
    &=
    \frac{1}{|x|},
    &
    j&\in\{0,\ldots,|x|-1\},
    \\
    q_{\tau,4}(*) &= 1.
\end{align*}
As before, when $|x|=0$, we use a singleton dummy parameter space for the
index-dependent schemas.

\smallsec{Character lists}
When $\tau=\texttt{List Char}$, the domain is
\[
D_\tau= \left(
\Sigma_{\mathrm{char}}^*
\triangleq
\bigcup_{n\in\mathbb{N}}
\Sigma_{\mathrm{char}}^n \right),
\]
where $\Sigma_{\mathrm{char}}$ denotes the finite set of allowable characters. We define
\[
\mathcal{O}_{\tau}
\triangleq
\left\{
s_{\tau,1},
s_{\tau,2},
s_{\tau,3},
s_{\tau,4}
\right\},
\]
corresponding to character modification, appending, deletion, and reversal:
\begin{align*}
    s_{\tau,1}(x;j,c)
    &=
    \begin{cases}
    \operatorname{replace}(x,j,c), & |x|>0,\\
    x, & |x|=0,
    \end{cases}
    \\
    s_{\tau,2}(x;c)
    &= x \mathbin{++} [c],
    \\
    s_{\tau,3}(x;j)
    &=
    \begin{cases}
    \operatorname{del}(x,j), & |x|>0,\\
    x, & |x|=0,
    \end{cases}
    \\
    s_{\tau,4}(x;*)
    &= \operatorname{rev}(x).
\end{align*}
The parameter spaces are
\begin{align*}
    \Theta_{\tau,1}(x)
    &=
    \{0,\ldots,|x|-1\}
    \times
    \Sigma_{\mathrm{char}},
    \\
    \Theta_{\tau,2}
    &=
    \Sigma_{\mathrm{char}},
    \\
    \Theta_{\tau,3}(x)
    &=
    \{0,\ldots,|x|-1\},
    \\
    \Theta_{\tau,4}
    &= \varnothing.
\end{align*}
When ingredient pool reuse is disabled, the schema is sampled uniformly:
\[
p_{\tau}(i)=\frac{1}{4},
\quad
i\in\{1,2,3,4\}.
\]
The parameter distributions are uniform:
\begin{align*}
    q_{\tau,1}(j,c\mid x)
    &=
    \frac{1}{|x|}\cdot\frac{1}{|\Sigma_{\mathrm{char}}|},
    &
    j&\in\{0,\ldots,|x|-1\},
    \quad
    c\in\Sigma_{\mathrm{char}},
    \\
    q_{\tau,2}(c)
    &=
    \frac{1}{|\Sigma_{\mathrm{char}}|},
    &
    c&\in\Sigma_{\mathrm{char}},
    \\
    q_{\tau,3}(j\mid x)
    &=
    \frac{1}{|x|},
    &
    j&\in\{0,\ldots,|x|-1\},
    \\
    q_{\tau,4}(*) &= 1.
\end{align*}
For $|x|=0$, the index-dependent schemas use a singleton dummy parameter space and reduce to the identity transformation.

\smallsec{Strings}
When $\tau=\texttt{String}$, we model the domain in the same way as
$\texttt{List Char}$, namely as the set of all finite character sequences:
\[
\Dd_\tau = \Sigma_{\mathrm{char}}^*.
\]
Unlike $\texttt{List Char}$, however, we define the mutation-schema set as
\[
\Oo_{\tau}
\triangleq
\left\{
s_{\tau,1},
s_{\tau,2},
s_{\tau,3}
\right\},
\]
where
\begin{align*}
    s_{\tau,1}(x)
    &= \epsilon,
    &
    \Theta_{\tau,1}
    &= \varnothing,
    \\
    s_{\tau,2}(x)
    &= \operatorname{rev}(x),
    &
    \Theta_{\tau,2}
    &= \varnothing,
    \\
    s_{\tau,3}(x;c)
    &= x \mathbin{++} c,
    &
    \Theta_{\tau,3}
    &= \Sigma_{\mathrm{sp}}.
\end{align*}
Here, $\epsilon$ denotes the empty string, $\operatorname{rev}$ reverses a
string, and $\Sigma_{\mathrm{sp}}$ is the set of special characters. When
ingredient pool reuse is disabled, the schema is sampled uniformly:
\[
p_{\tau}(i)=\frac{1}{3},
\quad
i\in\{1,2,3\}.
\]
The parameter distributions are
\begin{align*}
    q_{\tau,1}(*) &= 1,\\
    q_{\tau,2}(*) &= 1,\\
    q_{\tau,3}(c)
    &= \frac{1}{|\Sigma_{\mathrm{sp}}|},
    \quad c\in\Sigma_{\mathrm{sp}}.
\end{align*}


\section{Hyperparameter Settings}
\label{sec:hyperparameter}
Table~\ref{tab:hyperparams} provides the detailed hyperparameter configurations used across the test-suite expansion and reduction stages of the VeriScale pipeline.

\begin{table}[h]
\centering
\small
\setlength{\tabcolsep}{6pt}
\renewcommand{\arraystretch}{1.1}
\caption{Detailed hyperparameter settings for the VeriScale pipeline.}
\label{tab:hyperparams}
\begin{tabular}{lcp{8.5cm}}
\toprule[1pt]
\textbf{Parameter} & \textbf{Value} & \textbf{Description} \\
\midrule[0.8pt]
\multicolumn{3}{l}{\textbf{Test-Suite Expansion}} \\
\midrule[0.1pt]
\multicolumn{3}{l}{\textit{Seed Generation}} \\
\texttt{--rounds} & 1 & Number of LLM generation rounds for each task\\
\texttt{--candidates\_per\_round} & 40 & Number of candidates generated per round for each task\\
\texttt{--example\_limit} & 5 & Maximum number of accept/reject examples in prompt \\
\midrule[0.1pt]
\multicolumn{3}{l}{\textit{Type-aware Mutation}} \\
\texttt{--max\_mutations\_per\_input} & 15 & Upper limit of mutation samples per seed input \\
\texttt{--mutation\_multi\_step\_size} & 5 & Maximum number of consecutive steps per mutation \\
\texttt{--mutation\_ingredient\_prob} & 0.3 & Probability of ingredient reuse during mutation \\
\midrule[0.1pt]
\multicolumn{3}{l}{\textit{Adversarial Implementation Synthesis}} \\
\texttt{--max\_adver\_impl} & 5 & Maximum number of adversarial implementations per specification \\
\midrule[0.5pt]
\multicolumn{3}{l}{\textbf{Test-Suite Reduction}} \\
\midrule[0.1pt]
\textit{Boundary-Preserving Reduction} \\
\texttt{--MAX\_REJECT\_INPUTS\_PER\_TASK} & 50 & Maximum number of unexpected inputs retained per task \\
\texttt{--KEEP\_PER\_CRITICAL\_BUCKET} & 1 & Number of unexpected inputs retained from each boundary bucket \\
\midrule[0.1pt]
\textit{Adversary-Killing Reduction} \\
\texttt{--MAX\_ACCEPT\_TEST\_CASES\_PER\_TASK} & 50 & Maximum number of expected input-output pairs retained per task \\
\bottomrule[1pt]
\end{tabular}
\end{table}

\newpage
\section{Prompt Templates}
\label{sec:prompts}
This appendix provides the detailed prompt templates utilized across the expansion stage of the VeriScale pipeline. 
We present the exact instructions formulated for tasks including seed generation, problem decomposition, specification generation, and adversarial implementation synthesis.

\begin{tcolorbox}[
    breakable,
    colback=gray!5!white,
    colframe=black!70!white,
    title={Prompt Template for Seed Generation},
    fonttitle=\bfseries\small,
    boxrule=0.6pt,
    arc=1.5mm,
    left=2mm, right=2mm, top=2mm, bottom=2mm
]
\begin{lstlisting}[
    language={},
    basicstyle=\small,
    breaklines=true,
    breakindent=0pt,
    xleftmargin=0pt,
    columns=fullflexible,
    frame=none,
    keepspaces=true,
    showstringspaces=false,
    keywordstyle=\color{black},
    stringstyle=\color{black},
    commentstyle=\color{black},
    identifierstyle=\color{black}
]
You are an expert at generating diverse candidate inputs for Lean4 code verification tasks.

Return ONLY a JSON array. Each element must strictly follow:
{"input": {"param1": value1, "param2": value2, ...}}

Rules:
1. Do not output markdown, prose, comments, or code fences.
2. Keys in each `input` object must exactly match the function parameter names: no missing keys, no extra keys.
3. Values must be JSON-serializable and type-compatible with the declared Lean parameter types.
4. Treat the ground-truth precondition as the valid-input boundary: generate both valid and invalid inputs, especially boundary cases.
5. Include both likely-valid and likely-invalid inputs. Treat likely-invalid as semantically violating the precondition, not malformed JSON.
6. Generate diverse and challenging edge-case candidates. Return exactly the number of candidates requested by the user.

Generate candidate inputs for this Lean4 programming verification task.
Task description:
{description}

Ground-truth precondition:
{precond}

Use of precondition:
- It defines the semantic boundary of valid inputs.
- Generate both inputs that satisfy it and inputs that violate it.
- Focus on hard boundary/edge cases around this condition.
- "likely-invalid" means: still valid JSON and type-compatible, but likely violates the precondition.
Validity mix target:
- total candidates: {candidate_count}
- likely-invalid target: {invalid_target}
- likely-valid target: {valid_target}
- If constraints make exact ratio hard, prioritize exact total count and boundary coverage.

Function parameters (JSON):
{parameters}

Example likely-valid inputs:
{test_examples}

Example likely-invalid inputs (maybe no examples available, but generate if possible):
{reject_examples}

Output format MUST be exactly a JSON array of objects:
[{{"input": {{"param": value, ...}}}}, ...]
\end{lstlisting}
\end{tcolorbox}

\begin{tcolorbox}[
    colback=gray!5!white,
    colframe=black!70!white,
    title={Prompt Template for Problem Decomposition},
    fonttitle=\bfseries\small,
    boxrule=0.6pt,
    arc=1.5mm,
    left=2mm, right=2mm, top=2mm, bottom=2mm
]
\begin{lstlisting}[
    language={},
    basicstyle=\small,
    breaklines=true,
    breakindent=0pt,
    xleftmargin=0pt,
    columns=fullflexible,
    frame=none,
    keepspaces=true,
    showstringspaces=false,
    keywordstyle=\color{black},
    stringstyle=\color{black},
    commentstyle=\color{black},
    identifierstyle=\color{black}
]
### Role
You are a mathematical modeling expert. Your task is to analyze a programming problem description and produce a structured mathematical model that captures its formal meaning.

### Input
- **Problem Description**: A natural language description of a programming problem.

### Instructions
1. Identify the Input information: 
    a. contents (decleration of the input variables)
    b. constraints of the input
        - type/data structure of the input variables
        - properties of the input variables must satisfy, mark each property with one of the following two keywords 
            (E: Explicit, properties that are stated explicitly in the problem descripition)
            (H: Hypothesis, properties that may be included in the constraints)
2. Identify the Output information:
    a. contents (decleration of the output variables, you can declare them as `res1, res2, ...`)
    b. constraints of the output
        - type/data structure of the output variables
        - properties of the output variables must satisfy, the relationship between the input and the output (eg. elements/ordering preserving, elements inclusion, etc.) 
3. A detailed description of the type of input and output:
   - Figure out the data struture, consider list/tuples and etc.
   - Consider the type of the variables, for example, real numbers, integers, string and etc.  
4. A detailed description of the relationship between the input and the output. You may think through:
   - Does every element of the output originate from the input?
   - Does the output preserve any ordering from the input?
   - Does the output contain all elements from the input that meet some conditions?
5. Unless explicitly stated, do NOT mark non-empty, integer-only, positivity, uniqueness, sortedness and etc. with `E`.
6. Do not add any explanations of the constraints.

### Output Format
You MUST structure your response strictly as follows:

Input:
    a. {contents}
    b. {constraints}
        - {constraints1}
        - {constraints2}
        - ...
Output:
    a. {contents}
    b. {constraints}
        - {constraints1}
        - {constraints2}
        - ...

### Now perform the following task:
- **Problem Description**:
{problem_description}
\end{lstlisting}
\end{tcolorbox}

\begin{tcolorbox}[
    colback=gray!5!white,
    colframe=black!70!white,
    title={Prompt Template for Specification Generation},
    fonttitle=\bfseries\small,
    boxrule=0.6pt,
    arc=1.5mm,
    left=2mm, right=2mm, top=2mm, bottom=2mm
]
\begin{lstlisting}[
    language={},
    basicstyle=\small,
    breaklines=true,
    breakindent=0pt,
    xleftmargin=0pt,
    columns=fullflexible,
    frame=none,
    keepspaces=true,
    showstringspaces=false,
    keywordstyle=\color{black},
    stringstyle=\color{black},
    commentstyle=\color{black},
    identifierstyle=\color{black}
]
### Role
You are an expert in Lean4 programming. Your task is to analyze a programming problem description and produce the Lean4 code of its precondition and postcondition.

### Input
- **Problem Description**: A natural language description of a programming problem.
- **Input**: Input of the problem, and constraints on the input.
- **Output**: Output of the problem, and constraints on the output.
- **Precondition Function Signature**: Function signature of the precondition.
- **Postcondition Function Signature**: Function signature of the postcondition.

### Instructions
1. Generate the Lean4 code for the precondition and postcondition according to the Input and Output information. Both the Input and Output have two parts: contents and constraints. Note that you should include all the information provided for you.
2. You should carefully analyze the problem description to help you identify gaps and omissions, thereby capturing any missing details within the Input and Output.
3. Define the Lean4 precondition and postcondition by using the provided function signature.
4. Only translate the constraints of the precondition marked with 'E'. Neglect the constraints marked with 'H'.
5. If auxiliary functions are needed, put them directly before the precondition and postcondition.

### Output Format
You MUST structure your response strictly as follows:
```lean
-- Precondition Implementation
{precondition auxiliary functions, if any. precondition implementation}

-- Postcondition Implementation
{postcondition auxiliary functions, if any. postcondition implementation}
```

### Now perform the task below:
- **Problem Description**:
{problem_description}
- **Input**:
{input}
- **Output**:
{output}
- **Precondition Function Signature**:
{precond_signature}
- **Postcondition Function Signature**:
{postcond_signature}
\end{lstlisting}
\end{tcolorbox}

\newpage
\begin{tcolorbox}[
    breakable,
    colback=gray!5!white,
    colframe=black!70!white,
    title={Prompt Template for Adversarial Implementation Synthesis},
    fonttitle=\bfseries\small,
    boxrule=0.6pt,
    arc=1.5mm,
    left=2mm, right=2mm, top=2mm, bottom=2mm
]
\begin{lstlisting}[
    language={},
    basicstyle=\small,
    breaklines=true,
    breakindent=0pt,
    xleftmargin=0pt,
    columns=fullflexible,
    frame=none,
    keepspaces=true,
    showstringspaces=false,
    keywordstyle=\color{black},
    stringstyle=\color{black},
    commentstyle=\color{black},
    identifierstyle=\color{black},
    literate=
        {α}{{$\alpha$}}1
        {∃}{{$\exists$}}1
        {∀}{{$\forall$}}1
        {∈}{{$\in$}}1
]
### Role
You are an expert Red Teamer in formal verification and a "lazy" code hacker. Your objective is to perform "Specification Gaming": identify underspecification loopholes within a given `post-condition` in Lean4, and write degenerate code to exploit these vulnerabilities.

### Input
- **Problem Description**: A natural language description of a programming problem.
- **Pre-condition**: Defines valid inputs. Assume inputs always pass this.
- **Post-condition**: The flawed contract your output MUST satisfy.
- **Implementation Signatures**: The function signature of the adversarial functions.

### Instructions
1. You MUST NOT write a genuine algorithmic solution to the `Problem Description`. **If your code actually solves the true problem perfectly, you have FAILED your mission.** Your code MUST produce fundamentally WRONG outputs according to the natural language problem, BUT these wrong outputs must magically trick the `post-condition` into evaluating to `True` (or being provable).
2. Carefully analyze the `post-condition`. What crucial constraints from the `Problem Description` did it forget to check? You may consider
   - Did it check `List.length` but forget to check elements?
   - Did it check `List.Sorted` but forget to ensure elements belong to the original list?
   - Did it use `∃` without bounding the witness?
   - Did it use `∀ x ∈ xs` but forget the empty-list case is vacuously true?
3. Here are some adversarial strategies you may consider:
   - *Constant Return*: Return meaningless constants (`0`, `#[]`, `[]`, `""`, `#[0, 0]`, `false`, `none`) if the post-condition fails to bind the output to the input.
   - *Input Echoing*: Return the input parameters directly unmodified (e.g., `fun xs => xs`).
   - *Trivial Synthesis*: Return `List.replicate n 0`, `List.range n`, `Array.mkArray n 0`, or entirely fabricated data that satisfies the shape/type/length constraints.
   - *Malicious Deletion/Modification*: If the post-condition only checks for the absence of something (e.g., "no duplicates", "no negative elements"), delete everything and return `[]` / `#[]`.
   - *Edge Case Exploitation*: Force empty or trivial states (e.g., empty list, `0`, `none`) that make universally quantified checks (`∀ x ∈ []`, ...) vacuously true.
4. Your adversarial functions MUST be pure and total Lean4 functions. You MUST NOT:
   - declare new `axiom`s, use `sorry`, or introduce `opaque` definitions to bypass verification
   - use `unsafe`, `partial`, or non-terminating recursion to avoid producing a real value
   - use `@[implemented_by ...]`, `@[extern ...]`, or any attribute that replaces the compiled behavior of a function
   - register new `instance`s that shadow or override existing typeclass resolution for standard types (`Eq`, `LE`, `LT`, `Decidable`, `Ord`, `BEq`, `Hashable`, etc.)
   - rely on `IO`, `ST`, `IO.Ref`, or any monadic side effects that persist outside the function call
   - use noncomputable definitions or use `choose`, `choose_spec` to define the adversarial functions
   If any strategy requires modifying the ambient environment, DO NOT use it.
5. You MUST NOT use type-level / typeclass / coercion tricks to fake post-condition satisfaction. In particular, DO NOT:
   - define custom `structure`s / `inductive` types as outputs purely so that a custom `DecidableEq`, `LE`, `LT`, or `BEq` instance makes the post-condition trivially true
   - override or re-declare instances such as `BEq`, `DecidableEq`, `LE`, `LT`, `Ord`, `HAdd`, `HSub`, `Membership`, `GetElem` for standard types
   - return proxy/wrapper/sentinel values (e.g., via `Subtype` with a bogus proof, or a wrapper with a custom equality) that are not semantically valid outputs
   - exploit `Decidable` instances whose `decide` always returns `isTrue`
   Outputs must be plain values of the exact types expected by the specification (`Nat`, `Int`, `String`, `List α`, `Array α`, `Option α`, tuples of these, etc.).
6. You must generate exactly **5** different adversarial implementations. Try to use a *different* adversarial strategy from the list above for each one to probe different potential loopholes.
7. The function signatures of your adversarial implementations MUST match the implementation signature provided. Distinguish each adversarial function by appending the suffix `i` to its name, where i denotes the index of the i-th adversarial implementation.
8. DO NOT add any explanations.

### Output Format
You MUST structure your response strictly as follows. Do not include any explanations, and you must provide **5** adversarial implementations.

```lean
-- Adversarial Implementation 1
{adversatial function 1}

-- Adversarial Implementation 2
{adversatial function 2}

-- Adversarial Implementation 3
{adversatial function 3}

-- Adversarial Implementation 4
{adversatial function 4}

-- Adversarial Implementation 5
{adversatial function 5}
```

### Now perform the task below:
- **Problem Description**:
{problem_description}
- **Pre-condition**:
```lean
{pre_condition}
```
- **Post-condition**:
```lean
{post_condition}
```
- **Implementation Signatures**:
```lean
{impl_signature}
```
\end{lstlisting}
\end{tcolorbox}

\end{document}